\documentclass[12pt]{article}

\usepackage{sbc-template}
\usepackage{graphicx,url}
\usepackage[utf8]{inputenc}
\usepackage{array}
\usepackage{amsmath}
\usepackage{booktabs}
\usepackage{multirow}
\usepackage{color}
\usepackage{todonotes}

\title{Generating Sense Embeddings for Syntactic and Semantic Analogy for Portuguese}

\author{J\'{e}ssica Rodrigues da Silva\inst{1}, Helena de Medeiros Caseli\inst{1} }

\address{Federal University of São Carlos, Department of Computer Science
  \email{jsc.rodrigues@gmail.com,
  helenacaseli@ufscar.br}
}

\begin{document} 

\maketitle

\begin{abstract}
  Word embeddings are numerical vectors which can represent words or concepts in a low-dimensional continuous space. These vectors are able to capture useful syntactic and semantic information. The traditional approaches like Word2Vec, GloVe and FastText have a strict drawback: they produce a single vector representation per word ignoring the fact that ambiguous words can assume different meanings. In this paper we use techniques to generate sense embeddings and present the first experiments carried out for Portuguese. Our experiments show that sense vectors outperform traditional word vectors in syntactic and semantic analogy tasks, proving that the language resource generated here can improve the performance of NLP tasks in Portuguese.
\end{abstract}

\section{Introduction}
\label{intro}

Any natural language (Portuguese, English, German, etc.) has ambiguities. Due to ambiguity, the same word surface form can have two or more different meanings. For example, the Portuguese word \textit{banco} can be used to express the financial institution but also the place where we can rest our legs (a seat). Lexical ambiguities, which occur when a word has more than one possible meaning, directly impact tasks at the semantic level and solving them automatically is still a challenge in natural language processing (NLP) applications. One way to do this is through word embeddings.

Word embeddings are numerical vectors which can represent words or concepts in a low-dimensional continuous space, reducing the inherent sparsity of traditional vector-space representations \cite{salton1975vector}. These vectors are able to capture useful syntactic and semantic information, such as regularities in natural language. They are based on the distributional hypothesis, which establishes that the meaning of a word is given by its context of occurrence \cite{bruni2014multimodal}. The ability of embeddings to capture knowledge has been exploited in several tasks, such as Machine Translation \cite{mikolov2013exploiting}, Sentiment Analysis \cite{socher2013recursive}, Word Sense Disambiguation \cite{chen2014unified} and Language Understanding \cite{mesnil2013investigation}.

Although very useful in many applications, the word embeddings (word vectors), like those generated by Word2Vec \cite{mikolov2013Aefficient}, GloVe \cite{pennington2014glove} and FastText \cite{bojanowski2016enriching} have an important limitation: the Meaning Conflation Deficiency, which is the inability to discriminate among different meanings of a word. In any natural language, there are words with only one meaning (monosemous) and words with multiple meanings (ambiguous) \cite{CamachoCollados2018FromWT}. In word embeddings, each word is associated with only one vector representation ignoring the fact that ambiguous words can assume different meanings for which different vectors should be generated. Thus, there is a loss of information by representing a lexical ambiguity in a single vector, since it will only contain the most commonly used meaning for the word (or that which occurs in the corpus from which the word vectors were generated).

Several works \cite{pina2014simple,neelakantan2015efficient,wu2015sense,liu2015multi,huang2012improving,reisinger2010multi,iacobacci2015sensembed} have investigated the representation of word senses instead of word occurrences in what has been called \textit{sense embeddings} (sense vectors).

In this paper, we present the first experiments carried out to evaluate sense vectors for Portuguese. In section \ref{sec:relatedwork} we describe some of the approaches for generating sense vectors proposed in the literature. The approaches investigated in this paper are described in section \ref{sec:sswe}. The experiments carried out for evaluating sense vectors for Portuguese are described in section~\ref{sec:experiment}. Section~\ref{sec:conclusion} finishes this paper with some conclusions and proposals for future work.

\section{Related Work}
\label{sec:relatedwork}

\cite{shutze1998discrimination} was one of the first works to identify the meaning conflation deficiency of word vectors and to propose the induction of meanings through the clustering of contexts in which an ambiguous word occurs. Then, many other works followed these ideas.

One of the first works using neural network to investigate the generation of sense vectors was \cite{reisinger2010multi}. The approach proposed there is divided in two phases: pre-processing and training. In the pre-processing, firstly, the context of each target word is defined as the words to the left and to the right of that target word. Then, each possible context is represented by the weighted average of the vectors of the words that compose it. These context vectors are grouped and each centroid is selected to represent the sense of the cluster. Finally, each word of the corpus is labeled with the cluster with the closest meaning to its context. After this pre-processing phase, a neural network is trained from the labeled corpus, generating the sense vectors. The model was trained in two corpora, a Wikipedia dump in English and the third English edition of Gigaword corpus. The authors obtained a correlation of Spearman of around 62.5\% in WordSim-353 \cite{finkelstein2001placing}\footnote{WordSim-353 is a dataset with 353 pairs of English words for which similarity scores were set by humans on a scale of 1 to 10.}, for the Wikipedia and Gigaword corpus.

Another approach for generating sense vectors was \cite{huang2012improving}, which extends the \cite{reisinger2010multi}'s approach by incorporating a global context into the generation of word vectors. According to them, aggregating information from a larger context improves the quality of vector representations of ambiguous words that have more than one possible local context. To provide the vector representation of the global context, the proposed model uses all words in the document in which the target word occurs, incorporating this representation into the local context. The authors trained the model in a Wikipedia dump (from April 2010) in English with 2 million articles and 990 million tokens. The authors obtained a Spearman correlation of 65.7\% in the Stanford's Contextual Word Similarities (SCWS)\footnote{The SCWS is a dataset with 2,003 word pairs in sentential contexts.}, surpassing the baselines.

Based on \cite{huang2012improving}, \cite{neelakantan2015efficient} proposed the generation of sense vectors by performing a Skip-Gram adaptation of \cite{mikolov2013Aefficient}. In this approach, the identification of the senses occurs together with the training to generate the vectors, making the process efficient and scalable. This approach was the one chosen to be used in this paper and is explained in detail in the next section. The authors used the same corpus as \cite{huang2012improving} for training the sense vectors and obtained a Spearman correlation of 67.3 \% also in the SCWS, surpassing the baselines.

\cite{trask2015sense2vec} propose a different approach that uses a tagged corpus rather than a raw corpus for sense vectors generation. The authors annotated the corpus with part of speech (PoS) tags and that allowed the identification of ambiguous words from different classes. For example, this approach allow to distinguish between the noun \textit{livro} (book) and the verb \textit{livro} (free). After that they trained a word2vec (CBOW or Skip-Gram) model \cite{mikolov2013Aefficient} with the tagged corpus. The authors did not report results comparing their approach with baselines. In addition to the PoS tags, the authors also tested the ability of the method to disambiguate named entities and feelings, also labeling the corpus with these tags, before generating word embeddings. This approach was one of the chosen to be investigated in this paper and it will be explained in detail in the next section.

More recently, new proposals for language model generation like ELMo \cite{peters2018elmo}, OpenAI GPT \cite{radford2018openai} and BERT \cite{devlin2018bert} have begun to use more complex architectures to model context and capture the meanings of a word. The idea behind this language models is that each layer of the neural network is able to capture a different sense of the input word and generate dynamic vector representations, according to each input context. This idea of dynamic embeddings facilitates the use of these representations in downstream tasks. These architectures are complex and require very powerful hardware resources for training. The difference between sense vectors and language models like those lies in the architecture and in the way the trained model is used. Sense vectors are features that will be used for specific NLP tasks. On the other hand, the complex architecture of language models has both the neural networks that will create the language model and the NLP tasks, which can even share the same hyper-parameters (fine-tuning approach).

\section{Sense embeddings}
\label{sec:sswe}

In this paper, two approaches were used for sense vectors generation: the MSSG \cite{neelakantan2015efficient} and the Sense2Vec \cite{trask2015sense2vec}. Each one is explained in the next sections.

\subsection{Multiple-Sense Skip-Gram (MSSG)}

In \cite{neelakantan2015efficient}, two methods were proposed for generating sense vectors based on the original Skip-Gram model \cite{mikolov2013Aefficient}: MSSG (Multiple-Sense Skip-Gram) and NP-MSSG (Non-Parametric Multiple-Sense Skip-Gram). The main difference between them is that MSSG implements a fixed amount of possible meanings for each word while NP-MSSG does this as part of its learning process.

In both methods, the vector of the context is given by the weighted average of the vectors of the words that compose it. The context vectors are grouped and associated to the words of the corpus by approximation to their context. After predicting the sense, the gradient update is performed on the centroid of the cluster and the training continues. The training stops when vector representations have already been generated for all the words.

Different from the original skip-gram, its extensions, MSSG and NP-MSSG, learn multiple vectors for a given word. They were based on works such as \cite{huang2012improving} and \cite{reisinger2010multi}. In the MSSG model, each word $w \in W$ is associated to a global vector $v_g(w)$ and each sense of the word has a sense vector $v_s(w,k)(k=1,2,\cdots,K)$ and a context cluster with centroid $u(w,k)(k=1,2,\cdots,K)$. The $K$ sense vectors and the global vectors are of dimension $d$ and $K$ is a hyperparameter.

Considering the word $w_t$, its context $c_t=\{w_{t-R_t},\cdots,w_{t-1},w_{t+1},\cdots,w_{t+R_t}\}$ and the window size $R_t$, vector representation of the context is defined as the mean of the global vector representation of the words in the context. Let $v_{context}(c_t)=\frac{1}{2*R_t}\sum_{c \in c_t} v_g(c)$ be the vector representation of the context $c_t$. The global vectors of context words are used instead of their sense vectors to avoid the computational complexity associated with predicting the meanings of words in the context. It is possible, then, to predict the meaning of the word $w_t$, $s_t$, when it appears in the context $c_t$. 


The algorithm used for building the clusters is similar to k-means. The centroid of a cluster is the mean of the vector representations of all contexts that belong to this cluster and the cosine similarity is used to measure the similarity.

In MSSG, the probability ($P$) that the word $c$ is observed in the context of the word $w_t$ ($D=1$), given the sense and the probability that it is not observed ($D=0$), has the addition of $s_t$ (sense of $w_t$) in the formulas of the original Skip-gram (formula \ref{eq:likelehood_context} and \ref{eq:likelehood_notcontext}). The objective function ($J$) also considers $(w_t, s_t)$ instead of just $(w_t)$ (formula \ref{eq:objective_function}).

\begin{equation}
P(D = 1|v_s(w_t, s_t), v_g(c)) = \frac{1}{1 + e^{-v_s(w_t, s_t)^T v_g(c)}}
\label{eq:likelehood_context}
\end{equation}

\begin{equation}
\label{eq:likelehood_notcontext}
P(D = 0|v_s(w_t, s_t),v_g(c)) = 1 - P(D = 1|v_s(w_t, s_t),v_g(c))
\end{equation}

\begin{equation}
\begin{split}
\label{eq:objective_function}
J = \sum_{(w_t,c_t) \in D_+} \sum_{c \in c_t} log P(D=1|v_s(w_t, s_t),v_g(c)) + \\ 
\sum_{(w_t,c'_t) \in D_-} \sum_{c' \in c'_t} log P(D=0|v_s(w_t, s_t),v_g(c'))
\end{split}
\end{equation}

After predicting the meaning of the word $w_t$, MSSG updates the sense vector generated for the word $w_t(v_s(w_t,s_t))$, the global vector of context words and the global vector of noisy context words selected by chance. The centroid of the context cluster $s_t$ for the word $w_t(u(w_t,s_t))$ is updated when the context $c_t$ is added to the cluster $s_t$. 

In this paper, we choose to work with the MSSG fixing the amount of senses for each target word. We did that to allow a fair comparison with the second approach investigated here which a limited amount of meanings.

\subsection{Sense2Vec}

\cite{trask2015sense2vec} propose the generation of sense vectors from a corpus annotated with part-of-speech (PoS) tags, making it possible to identify ambiguous words from the amount of PoS tags they receive (for example, the noun \textit{livro} (book) in contrast with the verb \textit{livro} (free)). 


The authors suggest that annotating the corpus with PoS tags is a costless approach to identify the different context of ambiguous words with different PoS tags in each context. This approach makes it possible to create a meaningful representation for each use. The final step is to train a word2vec model (CBOW or Skip-Gram) \cite{mikolov2013Aefficient} with the tagged corpus, so that instead of predicting a word given neighboring words, it predicts a sense given the surrounding senses.

\cite{trask2015sense2vec} presents experiments demonstrating the effectiveness of the method for sentiment analysis and named entity recognition (NER). For sentiment analysis, sense2vec was trained with a corpus annotated with PoS tags and adjectives with feeling tags. The word ``bad'' was disambiguated between positive and negative sentiment. For the negative meaning, words like ``terrible'', ``horrible'' and ``awful'' appeared, while in the positive meaning there was present words like ``good'', ``wrong'' and ``funny'', indicating a more sarcastic sense of ``bad''.

In the NER task, sense2vec was trained with a corpus annotated with PoS and NER tags. For example, the NE ``Washington'' was disambiguated between the entity categories PERSON-NAME (person's name) and GPE (geolocation). In the PERSON-NAME category it was associated with words like ``George-Washington'', ``Henry-Knox'' and ``Philip-Schuyler'' while in the GPE category the word was associated with ``Washington-DC'', ``Seattle'' and ``Maryland''.

\section{Experiments and Results}
\label{sec:experiment}

In this section we present the first experiments carried out to evaluate sense vectors generated for Portuguese. As follows, we first describe the corpora used to generate sense vectors, then we present the network parameters used for training the models and, finally, we show the experiments carried out to evaluate the two approaches under investigation: MSSG and sense2vec.

\subsection{Training Corpora}

The corpora used for the training of sense vectors were the same as \cite{hartmann2017portuguese} which is composed of texts written in Brazilian Portuguese (PT-BR) and European Portuguese (PT-EU). Table~\ref{tab:corpusembeddings} summarizes the information about these corpora: name, amount of tokens and types and a briefly description of the genre. 

\begin{table}[!ht]
  \centering
  \footnotesize
  \caption{Statistics of our training corpora}
    \begin{tabular}[t]{lrrl}
    \midrule\midrule
     Corpus&
     Tokens&
     Types&
     Genre\\
    \midrule
    LX-Corpus \cite{rodrigues2016lx} & 714,286,638 & 2,605,393 & Mixed genres\\
    Wikipedia & 219,293,003 & 1,758,191 &  Encyclopedic\\
    GoogleNews & 160,396,456 & 664,320 & Informative\\
    SubIMDB-PT & 129,975,149 & 500,302 & Spoken language\\
    G1 & 105,341,070 & 392,635 & Informative\\
    PLN-Br & 31,196,395 & 259,762 & Informative\\
    Literacy works of\newline public domain & 23,750,521 & 381,697 & Prose\\
    Lacio-web \cite{aluisio2003lacioweb} & 8,962,718 & 196,077 & Mixed genres\\
    Portuguese e-books & 1,299,008 & 66,706 & Prose\\
    Mundo Estranho & 1,047,108 & 55,000 & Informative\\
    CHC & 941,032 & 36,522 & Informative\\
    FAPESP & 499,008 & 31,746 & Science\\
    Textbooks & 96,209 & 11,597 & Didactic\\
    Folhinha & 73,575 & 9,207 & Informative\\
    NILC subcorpus & 32,868 & 4,064 & Informative\\
    Para Seu Filho Ler & 21,224 & 3,942 & Informative\\
    SARESP & 13,308 & 3,293 &  Didactic\\
    \textbf{Total} & \textbf{1,395,926,282} & \textbf{3,827,725}\\
    \midrule\midrule
    \end{tabular}
    \vspace{-1\baselineskip}
  \label{tab:corpusembeddings}
\end{table}

The corpora were pre-processed in order to reduce the vocabulary size. For the sense2vec model, the corpora were also PoS-tagged using the nlpnet tool \cite{fonseca2013twostep}, which is considered the state-of-art in PoS-tagging for PT-BR.



It is important to say that both approaches for generating sense vectors were trained with these corpora. The only difference is that the input for the MSSG is the sentence without any PoS tag while the input for the sense2vec is the sentence annotated with PoS tags.

\subsection{Network Parameters}

For all training, including baselines, we generated vectors of 300 dimensions, using the Skip-Gram model, with context window of five words. The learning rate was set to 0.025 and the minimum frequency for each word was set to 10. For the MSSG approach, the maximum number of senses per word was set to 3.

\subsection{Evaluation}



Based on \cite{hartmann2017portuguese}, this experiment is a task of syntactic and semantic analogies where the use of sense vectors is evaluated. Word vectors were chosen as baselines. 

\paragraph{\textbf{Dataset.}} The dataset of Syntactic and Semantic Analogies of \cite{rodrigues2016lx} has analogies in Brazilian (PT-BR) and European (PT-EU) Portuguese. In syntactic analogies, we have the following categories: adjective-to-adverb, opposite, comparative, superlative, present-participle, nationality-adjective, past-tense, plural, and plural-verbs. In semantic analogies, we have the following categories: capital-common-countries, capital-world, currency, city-in-state and family. In each category, we have examples of analogies with four words:

\paragraph{\textbf{adjective-to-adverb:}}
\begin{itemize}
    \item \textit{fantástico fantasticamente aparente aparentemente}
    \textbf{(syntactic)}\\
    fantastic fantastically apparent apparently
\end{itemize}
\vspace*{-0.5cm}
\paragraph{\textbf{capital-common-countries:}}
\begin{itemize}
    \item \textit{Berlim Alemanha Lisboa Portugal }\textbf{(semantic)} \\
    Berlin Germany Lisbon Portugal
\end{itemize}

\paragraph{\textbf{Algorithm.}} The algorithm receives the first three words of the analogy and aims to predict the fourth. Thus, for instance considering the previous example, the algorithm would receive Berlin (a), Germany (b) and Lisbon (c) and should predict Portugal (d). Internally, the following algebraic operation is performed between vectors:
\begin{equation}
    v (b) + v (c) - v (a) = v (d)
\end{equation}

\paragraph{\textbf{Evaluation metrics.}} The metric used in this case is accuracy, which calculates the percentage of correctly labeled words in relation to the total amount of words in the dataset.

\paragraph{\textbf{Discussion of results.}} Table \ref{tab:evaluation} shows the accuracy values obtained for the syntactic and semantic analogies. The Word2vec, GloVe and FastText were adopted as word vectors baselines since they performed well in \cite{hartmann2017portuguese} experiments. Note that the sense vectors generated by our sense2vec model outperform the baselines at the syntactic and semantic levels.

\begin{table}[!htb]
    \centering
    \footnotesize
    \caption{Accuracy values for the syntactic and semantic analogies}
    \begin{tabular}[t]{lccc|ccc}
        \midrule\midrule
        \multicolumn{1}{c}{\multirow{2}{*}{\textbf{Embedding}}} &
        \multicolumn{3}{c}{\textbf{PT-BR}} & \multicolumn{3}{c}{\textbf{PT-EU}}\\
        \cmidrule{2-7}
        \multicolumn{1}{c}{} & \textbf{Syntactic} & \textbf{Semantic} & \textbf{All} & \textbf{Syntactic} & \textbf{Semantic} & \textbf{All}\\
        \midrule
        Word2Vec (word) & 49.4 & 42.5 & 45.9 & 49.5 & 38.9 & 44.3 \\
        \midrule
        GloVe (word) & 34.7 & 36.7 & 35.7 & 34.9 & 34.0 & 34.4 \\
        \midrule
        FastText (word) & 39.9 & 8.0 & 24.0 & 39.9 & 7.6 & 23.9 \\
        \midrule
        MSSG (sense) & 23.0 & 6.6 & 14.9 & 23.0 & 6.3 & 14.7 \\
        \midrule
        Sense2Vec (sense) & \textbf{52.4} & \textbf{42.6} & \textbf{47.6} & \textbf{52.6} & \textbf{39.5} & \textbf{46.2} \\
        \midrule\midrule
    \end{tabular}
    \vspace{-1\baselineskip}
    \label{tab:evaluation}
\end{table}

 In syntactic analogies, the sense vectors generated by sense2vec outperform the word vectors generated by word2vec in opposite, nationality-adjective, past-tense, plural and plural-verbs. An example is shown in table \ref{tab:syntactic}. We can explain this type of success through an algebraic operation of vectors. When calculating v(\textit{aparentemente} (apparently)) + v(\textit{completo} (complete)) - v(\textit{aparente} (apparent)) the resulting vector of word2vec is v(\textit{incompleto} (incomplete)) when it should be v(\textit{completamente} (completely)). The correct option appears as the second nearest neighbor. 
 
 So, we can conclude that the sense2vec's PoS tag functions as an extra feature in the training of sense vectors, generating more accurate numerical vectors, allowing the correct result to be obtained. 
 
 \begin{table}[!htb]
  \centering
  \caption{Example of syntactic analogy predicted by word2vec and sense2vec}
  \begin{minipage}{\textwidth}
  \scalebox{.90}{
    \begin{tabular}[t]{l|l}
    \midrule\midrule
    word2vec & aparente aparentemente completo : completamente \textbf{(expected)}\\
    & aparente aparentemente completo : incompleto \textbf{(predicted)}\\
    \midrule
    sense2vec & aparente$|$ADJ aparentemente$|$ADV completo$|$ADJ : completamente$|$ADV \textbf{(expected)}\\
    & aparente$|$ADJ aparentemente$|$ADV completo$|$ADJ : completamente$|$ADV \textbf{(predicted)}\\
    \midrule\midrule
    \end{tabular}}
  \end{minipage}
  \label{tab:syntactic}
\end{table}

In semantic analogies, the sense vectors generated by sense2vec outperform the word vectors generated by word2vec in capital-world, currency and city-in-state. Examples of city-in-state are shown in table \ref{tab:semantic}.

\begin{table}[!htb]
  \begin{center}
  \footnotesize
  \caption{Example of semantic analogies predicted by word2vec and sense2vec}
    \begin{tabular}[t]{l|l}
    \midrule\midrule
    word2vec & arlington texas akron : kansas \textbf{(predicted)} ohio \textbf{(expected)}\\
    sense2vec & arlington$|$N texas$|$N akron$|$N : ohio$|$N \textbf{(predicted)(expected)}\\
    \midrule
    word2vec & bakersfield califórnia madison : pensilvânia \textbf{(predicted)} wisconsin \textbf{(expected)}\\
    sense2vec & bakersfield$|$N califórnia$|$N madison$|$N : wisconsin$|$N \textbf{(predicted)(expected)}\\
    \midrule
    word2vec & worcester massachusetts miami : seattle \textbf{(predicted)} flórida \textbf{(expected)}\\
    sense2vec & worcester$|$N massachusetts$|$N miami$|$N : flórida$|$N \textbf{(predicted)(expected)}\\
    \midrule\midrule
    \end{tabular}
    \vspace{-1\baselineskip}
  \label{tab:semantic}
  \end{center}
\end{table}

In this case, the PoS tag is always the same for all words: N (noun). This indicates that the success of sense2vec is related to the quality of sense vectors as a whole. As all words are tagged, this feature ends up improving the inference of all vector spaces during training.


Based on \cite{mikolov2013Aefficient}, who performs the algebraic operation: vector("King") - vector("Man") + vector("Woman") = vector("Queen"), this experiment explores the ability of sense vectors to infer new semantic information through algebraic operations.

\paragraph{\textbf{Dataset.}} The CSTNews dataset \cite{cardoso2011cstnews} contains 50 collections of journalistic documents (PT-BR) with 466 nouns annotated with meanings, from the synsets of wordnet. Therefore, 77\% of the nouns are ambiguous (with more than two meanings). Some ambiguous words were chosen for algebraic operations between vectors.

\paragraph{\textbf{Algorithm.}} The algorithm receives the first three words of the analogy and aims to predict the fourth. Thus, for instance considering the previous example, the algorithm would receive Man (a), Woman (b) and King (c) and should predict Queen (d). Internally, the following algebraic operation is performed between vectors:
\begin{equation}
    v (b) + v (c) - v (a) = v (d)
\end{equation}

\paragraph{\textbf{Evaluation metrics.}} This evaluation shows qualitative results, so a metric is not used.

\paragraph{\textbf{Discussion of results.}} To illustrate how sense vectors capture the meaning differences better than word vectors do, examples of algebraic operations using word vectors (generated by word2vec) and sense vectors (generated by MSSG) are shown below. 

This first example is for the ambiguous word \textit{banco} (bank) which has three predominant meanings: (1) reserve bank (soccer, basketball), (2) physical storage (trunk, luggage rack) and (3) financial institution (Santander, Pactual).

\begin{figure}[htb!]
  \includegraphics[width=0.9\textwidth]{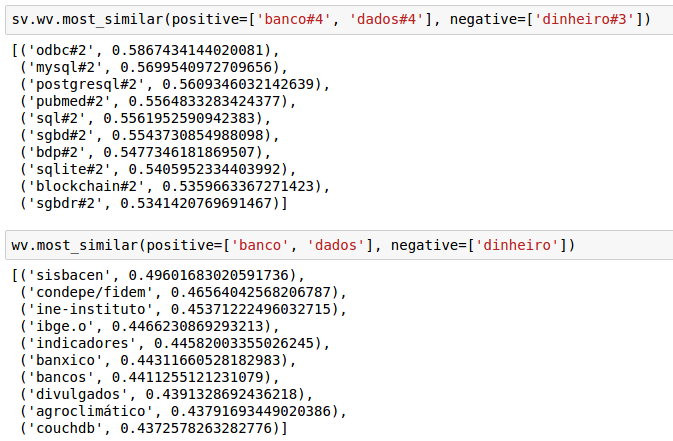}
\caption{Algebraic Operation by MSSG and word2vec with the word "banco"}
\label{fig:nearest_neighbor_bank}
\end{figure}

In this example, we show the results of \textit{banco} $+$ \textit{dados} $-$ \textit{dinheiro} (bank $+$ data $-$ money) and we expect as a result words related to the second meaning of the word \textit{banco}.\footnote{In Portuguese, the usual translation of ``database'' is \textit{banco de dados}. So, in Portuguese, MySQL, SQL, etc. are common words related to \textit{banco de dados}.} When the sense vectors are used (top part of Fig \ref{fig:nearest_neighbor_bank}) we obtain exactly what we were expecting. However, when the word vectors are used (bottom part of Fig \ref{fig:nearest_neighbor_bank}) we do not obtain any result related to data.

\begin{figure}[htb!]
  \includegraphics[width=0.9\textwidth]{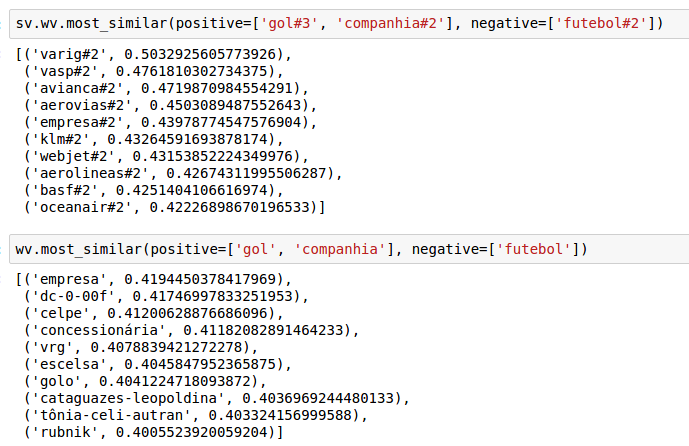}
\caption{Algebraic Operation by MSSG and word2vec with the word "gol"}
\label{fig:nearest_neighbor_gol}
\end{figure}

Another example is shown in Fig \ref{fig:nearest_neighbor_gol}. This example is for the ambiguous word \textit{gol} (goal) which has one predominant meanings: soccer goal. In this example, we show the results of \textit{gol} $+$ \textit{companhia} $-$ \textit{futebol} (goal $+$ company $-$ soccer) and we have discovered a new meaning for the word \textit{gol}: airline name such as KLM, LATAM and American Airlines (top part of Fig \ref{fig:nearest_neighbor_gol}). When the word vectors are used (bottom part of Fig \ref{fig:nearest_neighbor_gol}) we do not get this new meaning. With this algebraic operation, we can conclude that it is possible to discover new meanings for a word, even if it does not have a sense vector corresponding to this meaning.

The last example uses two ambiguous words in the same operation: \textit{centro} (center) and \textit{pesquisas} (researches). We have found interesting results for this operations. The ambiguous word \textit{centro} (center) has two predominant meanings: (1) institute (center of predictions, NASA center) and (2) midtown (central area). The ambiguous word \textit{pesquisas} (researches) which has two predominant meanings to: (1) scientific research (experiments, discoveries) and (2) opinion or market research.

\begin{figure}[htb!]
  \includegraphics[width=0.9\textwidth]{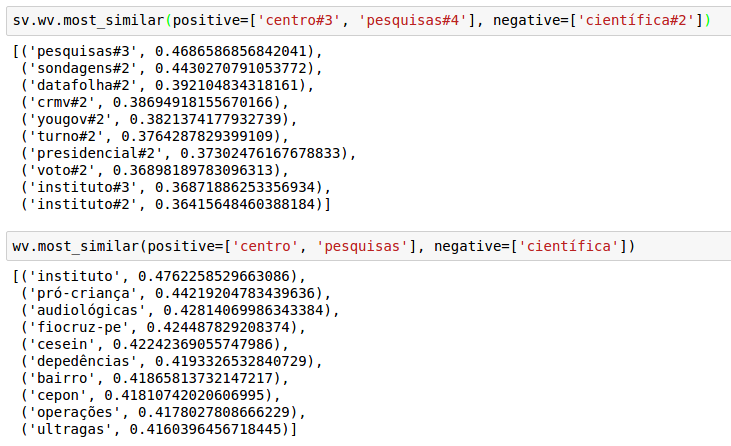}
\caption{Algebraic Operation by MSSG and word2vec with the word "centro$_{sense1}$"}
\label{fig:nearest_neighbor_centro_instituto}
\end{figure}

In the first operation, we show the results of \textit{centro$_{sense1}$} $+$ \textit{pesquisas$_{sense2}$} $-$ \textit{científica} (center$_{sense1}$ $+$ researches$_{sense2}$ $-$ scientific). In top part of figure \ref{fig:nearest_neighbor_centro_instituto}, we obtain a new type: institutes conducting statistical surveys, like Datafolha and YouGov, next to words related to the elections.

\begin{figure}[htb!]
  \includegraphics[width=0.9\textwidth]{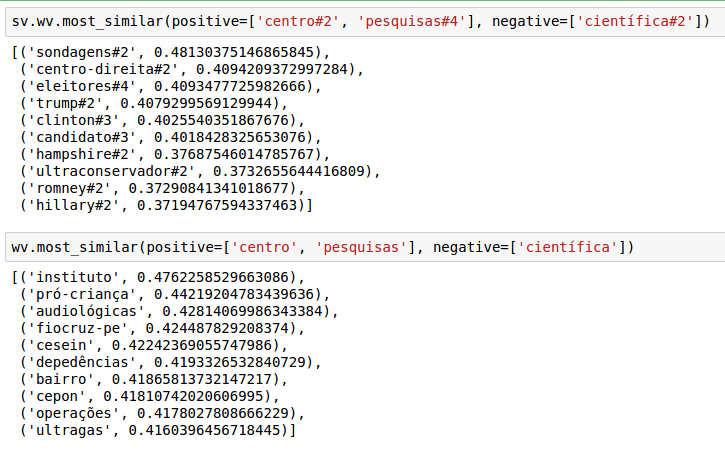}
\caption{Algebraic Operation by MSSG and word2vec with the word "centro$_{sense2}$"}
\label{fig:nearest_neighbor_centro_meio}
\end{figure}

In the second operation, we show the results of \textit{centro$_{sense2}$} $+$ \textit{pesquisas$_{sense2}$} $-$ \textit{científica} (center$_{sense2}$ $+$ researches$_{sense2}$ $-$ scientific). In top part of figure \ref{fig:nearest_neighbor_centro_meio}, we obtain a new type to: Political ideologies/orientation of left, right or \textbf{center}, with words like trump (Donald Trump), clinton (Bill Clinton), romney (Mitt Romney), hillary (Hillary Clinton) and also words like \textit{centro-direita} (center-right), \textit{ultraconservador} (ultraconservative), \textit{eleitores} (voters) and \textit{candidato} (candidate). These words are related to the political sides, including the names of right, left and \textbf{center} politicians.

These results are interesting because they show new nuances of the meanings and prove that it is possible to infer new semantic information, which is not represented as sense vectors. These findings are not made using word vectors (bottom part of Fig \ref{fig:nearest_neighbor_centro_instituto} and \ref{fig:nearest_neighbor_centro_meio}).



\section{Conclusion and Future Work}
\label{sec:conclusion}

In this paper we used techniques to generate sense embeddings (sense vectors) for Portuguese (Brazilian and European). The generated models were evaluated through the task of syntactic and semantic analogies and the accuracy values show that the sense vectors (sense2vec) outperform the baselines of traditional word vectors (word2vec, Glove, FastText) with a similar computational cost.

Our sense-vectors and the code used in all the experiments presented in this paper are available at \url{https://github.com/LALIC-UFSCar/sense-vectors-analogies-pt}. The application of sense vectors in NLP tasks (WSD and others) is under development.
As future work we intend to experiment a combination of the two approaches (MSSG and sense2vec) and also to explore how the new approaches proposed for generating language models perform in Portuguese.

\section*{Acknowledgements}
\label{sec:acknowledgements}

This research is part of the MMeaning project, supported by São Paulo Research Foundation (FAPESP), grant \#2016/13002-0, and was also partly funded by the Coordenação de Aperfeiçoamento de Pessoal de Nível Superior - Brasil (CAPES) - Funding Code 001.

\bibliographystyle{sbc}
\bibliography{sbc-template}

\end{document}